\documentclass{article} 
\usepackage{iclr2027_conference,times}

\usepackage{amsmath,amsfonts,bm}

\def\eqref#1{equation~\ref{#1}}

\def\1{\bm{1}}

\DeclareMathAlphabet{\mathsfit}{\encodingdefault}{\sfdefault}{m}{sl}
\SetMathAlphabet{\mathsfit}{bold}{\encodingdefault}{\sfdefault}{bx}{n}

\usepackage{booktabs}
\usepackage{xcolor}
\usepackage{threeparttable}
\usepackage{algorithm}
\usepackage{algpseudocode}
\usepackage{hyperref}
\usepackage{url}
\usepackage{latexsym}
\usepackage[T1]{fontenc}
\usepackage[utf8]{inputenc}
\usepackage{microtype}
\usepackage{inconsolata}
\usepackage{graphicx}
\usepackage{booktabs}
\usepackage{multirow}
\usepackage{array}
\usepackage{xspace}
\usepackage{fontawesome5}
\usepackage{booktabs}
\usepackage{multirow}
\usepackage{makecell}
\title{FACET: Preserving Source Intent and Executable State in Terminal Task Synthesis}

\author{
    \small
    \textbf{Kou Shi\textsuperscript{1}},
    \textbf{Zun Wang\textsuperscript{2}},
    \textbf{Qisheng Su\textsuperscript{1,2}},
    \textbf{Shiting Huang\textsuperscript{1}},
    \textbf{Ziao Zhang\textsuperscript{1}},
    \textbf{Zhen Fang\textsuperscript{1}},
    \textbf{Qingnan Ren\textsuperscript{1}}
    \\
    \small
    \textbf{Jin Liu\textsuperscript{3}},
    \textbf{Yu Zeng\textsuperscript{1}},
    \textbf{Yiming Zhao\textsuperscript{1}},
    \textbf{Lin Chen\textsuperscript{1}},
    \textbf{Zehui Chen\textsuperscript{1}},
    \textbf{Feng Zhao\textsuperscript{1,*}}
    \\[0.35em]
    \textsuperscript{1}MoE Key Lab of BIPC, University of Science and Technology of China
    \\
    \textsuperscript{2}Shanghai AI Laboratory,
    \qquad
    \textsuperscript{3}Fudan University
    \\[0.25em]
    \small
    \textbf{Contact:}
    \href{mailto:stokou@mail.ustc.edu.cn}{stokou@mail.ustc.edu.cn}
    \\
    \small
    \textbf{\textsuperscript{*}Correspondence:}
    \href{mailto:fzhao956@ustc.edu.cn}{fzhao956@ustc.edu.cn}
}

\newcommand{\benchname}{\textsc{FACET}\xspace}

\iclrfinalcopy
\begin{document}

\maketitle

\begin{center}
\vspace{-0.4cm} 
    \urlstyle{same}
    \href{https://github.com/StoKou/FACET-Terminal}
        {\faGithub\ \textbf{Code}}
    \qquad
    \href{https://huggingface.co/FACET-Terminal}
        {\faRobot\ \textbf{Model \& Datasets}}
    \qquad
    \href{https://stokou.github.io/FACET-Terminal/}
        {\faGlobe\ \textbf{Project Page}}
\end{center}
\vspace{0.4cm}

\begin{abstract}
Training terminal agents requires scalable executable supervision, yet
synthesizing high-quality terminal tasks remains challenging. Each task couples
an instruction, an initialized environment, a reference solution, and an
executable verifier; if these artifacts are generated from inconsistent
assumptions, the resulting task may be unsolvable or incorrectly evaluated.
Meanwhile, multi-stage synthesis can discard the goals, dependencies, state
transitions, and procedural constraints encoded in the original sources. We
present FACET (\textbf{F}ine-grained \textbf{A}gentic \textbf{C}onstruction of
\textbf{E}xecutable \textbf{T}asks), a framework that addresses both information
preservation and cross-artifact consistency. FACET reconstructs related agent
skills into coherent, information-rich scenarios, then realizes and repairs the
execution environment before generating the final task artifacts. The resulting
container state serves as shared grounding for the instruction, solution, and
verifier, while execution-based validation and targeted repair correct
artifact-specific failures without unnecessarily regenerating valid components.
FACET produces complex terminal tasks with dense executable checks, and
successful trajectories collected from these tasks provide effective,
data-efficient supervision. Fine-tuning models across multiple scales
consistently improves performance on Terminal-Bench 2.1, while analyses of
alternative generation schemes support the importance of environment-grounded
construction for task validity and solution--verifier alignment. These results
establish source-intent preservation and shared executable-state grounding as
key principles for scalable terminal-task synthesis.
\end{abstract}
\section{Introduction}
\label{sec:introduction}

A central goal in the pursuit of artificial general intelligence is to develop
systems that can not only reason about complex goals, but also act autonomously
and adapt in open-ended environments. Recent advances in language models have
shifted this pursuit beyond passive text generation toward agents that interact
with external tools and computer systems~\citep{openai2025introducingcodex,
anthropic2025claudecodebestpractices,mullen2025geminicli,
dohmke2025copilotcodingagent,openai2026swebenchverified}. To operate effectively in these environments,
agents must manipulate files, install dependencies, invoke command-line tools,
recover from execution failures, and complete long-horizon workflows
\citep{openai2025gpt52codex,anthropic2026claudecodepractice,
nvidia2026nemotronultra,metr2025longtasks}. Terminal environments provide a
natural interface for studying these capabilities because success depends not
only on generating plausible text or code, but also on interacting correctly
with a changing execution state~\citep{anthropic2025contextengineering,
anthropic2025longrunningharnesses,openai2026agentloop,
openai2026harnessengineering}. Benchmarks such as
Terminal-Bench~\citep{merrill2026terminalbench,terminalbench2026v21}
have therefore become important testbeds for evaluating agentic capabilities,
while recent work further shows that executable terminal tasks can provide
effective supervision for post-training language agents
\citep{gandhi2026endlessterminals,pi2026dataengineering,
peng2026litecoderterminal,ivison2026tmax}.

The growing demand for such supervision has motivated a range of synthetic
terminal-data pipelines. Existing approaches construct tasks from domain
specifications, skill taxonomies, reusable repositories
\citep{anthropic2025agentskills}, terminal recordings, community Q\&A, and
procedurally generated task signatures~\citep{fan2026skillsynth,
chu2026terminalworld,peng2026litecoderterminal,pi2026dataengineering,
gandhi2026endlessterminals,yang2026terminallego,zhao2026nexforge,
ivison2026tmax}. These efforts demonstrate the potential of synthetic
executable environments for scaling terminal-agent training. However,
increasing the number and diversity of source materials does not by itself
guarantee high-quality executable tasks. A terminal task is a tightly coupled
bundle of an instruction, an initialized environment, a reference solution,
and an executable verifier~\citep{harbor2026framework}; inconsistencies among
any of these components can render the entire task invalid.

We identify two challenges that become particularly important in multi-stage
terminal-task synthesis. First, \emph{source information is easily lost during
generation}. Rich source materials may contain capabilities, dependencies,
intermediate states, input--output contracts, and procedural constraints, yet
successive generation stages can progressively compress this information into
a simplified task description. Consequently, the synthesized task may preserve
only a fraction of the structure and complexity available in the original
sources. Second, \emph{task artifacts can drift apart during generation}.
An instruction may refer to a file that is not realized in the environment, a
solution may assume a different schema or dependency, or a verifier may test a
state that the generated task cannot produce. Passing textual specifications
between generation stages can mitigate this problem, but does not ensure that
all artifacts are grounded in the same realized execution state.

To address these challenges, we introduce \benchname, a framework for
synthesizing verifiable terminal tasks from large collections of reusable agent
skills. Rather than directly translating sampled skills into task descriptions,
\benchname first performs \emph{agentic scenario reconstruction} to recover
coherent user scenarios, cross-skill dependencies, intermediate states, and
solution workflows while retaining information from the original sources. It
then constructs and repairs the execution environment before finalizing the
task artifacts. The realized container state is exposed as a shared grounding
interface, allowing the instruction, solution, and verifier to be generated
against the same files, schemas, services, dependencies, and runtime state.
Finally, execution-based validation and targeted repair identify and correct
artifact-level failures while preserving components that are already valid.

Our main contributions are threefold:

\begin{itemize}
\item We introduce \benchname, a framework for synthesizing complex and
verifiable terminal tasks from heterogeneous agent skills. Rather than
treating skills as isolated task templates, our framework reconstructs
their underlying scenarios and workflows to better preserve the richness
of the source information during synthesis.

\item We propose an executable-state-grounded construction paradigm that
coordinates task artifacts through a shared, realized environment. This
shifts terminal-task synthesis from independently generating plausible
components toward constructing a coherent executable task as a whole.

\item We build a complete synthesis and validation pipeline for producing
reliable terminal-agent supervision. Extensive analyses characterize the
effects of generation design on task quality, while post-training
experiments across multiple model scales demonstrate the effectiveness of
the resulting data.

\end{itemize}

\section{\benchname}
\label{sec:pipeline}

\subsection{Problem Formulation}
\label{sec:problem-formulation}

Let $X=\{x_1,\ldots,x_m\}$ denote a set of related agent skills.
Our goal is to synthesize a terminal task bundle

\begin{equation}
    \mathcal{T}=(\mathcal{I},\mathcal{E},\mathcal{S},\mathcal{V},\mathcal{M}),
\end{equation}

\noindent where $\mathcal{I}$ is the user instruction, $\mathcal{E}$ is the
environment specification, $\mathcal{S}$ is the reference solution,
$\mathcal{V}$ is the executable verifier, and $\mathcal{M}$ contains the
runtime metadata.

Initializing $\mathcal{E}$ produces the initial state
$e_0=\operatorname{Init}(\mathcal{E})$, while executing $\mathcal{S}$ produces
$e_T=\operatorname{Run}(\mathcal{S},e_0)$ or failure $\bot$. Let
$B(\mathcal{E})$ indicate whether the environment builds successfully and
$\nu_{\mathcal{V}}(e)$ denote the verifier outcome on state $e$. A synthesized
task is accepted when

\begin{equation}
\mathcal{A}(\mathcal{T}) =
B(\mathcal{E})
\land \neg \nu_{\mathcal{V}}(e_0)
\land (e_T \neq \bot)
\land \nu_{\mathcal{V}}(e_T).
\end{equation}

\noindent This criterion requires a buildable environment, a non-trivial
initial state, an executable reference solution, and a verifier-accepted final
state.

\subsection{Information Source Acquisition}
\label{sec:source-processing}

\paragraph{Collection and filtering.}
We collect publicly available skill packages from
OpenClaw~\citep{openclaw2026}, ClawHub~\citep{clawhub2026}, and GitHub. We remove
skills containing unsafe instructions, requiring access to private websites or
information, or depending on non-public resources. Unreadable, non-actionable,
and duplicate records are also discarded.

\paragraph{Skill understanding.}
Each retained skill is normalized into a structured record containing its
description, required tools, inputs and outputs, procedural steps, and source
provenance. This process yields more than 71K valid skills, with detailed
statistics reported in Appendix~\ref{app:data}.

\paragraph{Scenario extraction.}
For each skill, an extraction agent identifies possible application contexts,
user goals, initial states, and desired final states. We embed these scenario
hypotheses and retrieve similar hypotheses from other skills to identify
potentially related skill combinations.

\paragraph{Scenario--skill repository construction.}
Similar hypotheses are grouped to construct candidate scenario--skill pairs
$p_c=(c,X_c)$. A model-based judge evaluates whether each scenario and its
associated skills are relevant, complementary, non-redundant, and executable as
a terminal workflow. Only candidates accepted by the judge are retained:
\begin{equation}
    \mathcal{P}
    =
    \left\{
        p_c \mid J(p_c)=1
    \right\},
    \label{eq:scenario-skill-pairs}
\end{equation}
where $\mathcal{P}$ is the final scenario--skill repository used by the
subsequent reconstruction stage. For analysis, we organize the retained skills
into five top-level and 34 fine-grained categories, and group the final
validated tasks into nine task families; detailed distributions are provided
in Appendix~\ref{app:data} and Figure~\ref{fig:distributions}.

\begin{figure*}[t]
    \centering
    \includegraphics[width=0.98\textwidth]{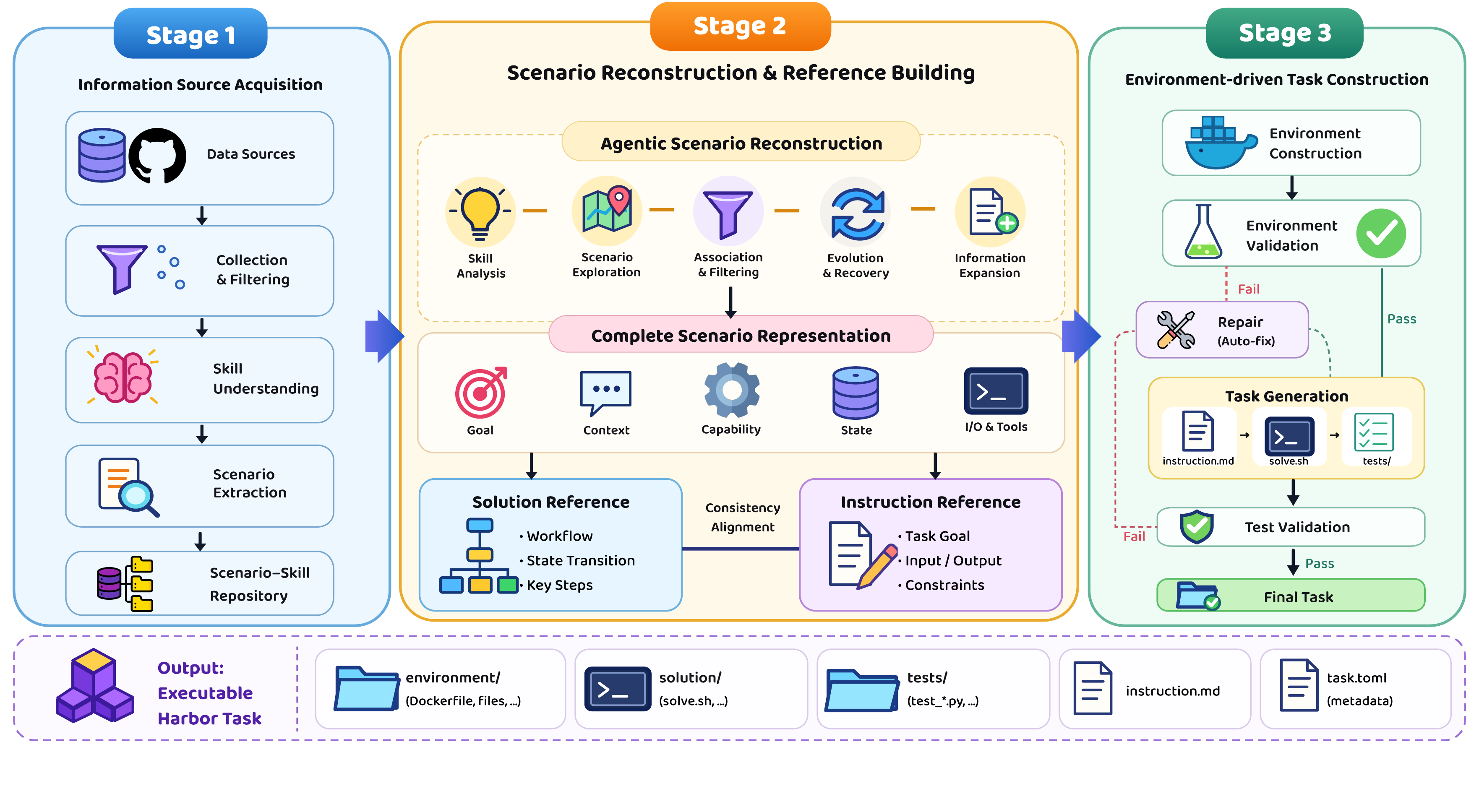}
    \caption{Overview of \benchname. Stage~1 collects skills and constructs the
    scenario--skill repository. Stage~2 understands the selected skills,
    explores and recovers a joint scenario, expands it into a complete
    representation, and builds aligned solution and instruction references.
    Stage~3 constructs and validates the environment before generating the
    final task artifacts, with bounded repair loops for failed builds and tests.}
    \label{fig:pipeline}
\end{figure*}

\subsection{Scenario Reconstruction and Reference Building}
\label{sec:reconstruction}

A scenario--skill pair provides sufficient information for task generation,
but directly converting it into final task artifacts can compress the source
information too early. This often produces shallow workflows that use only the
most apparent capability of each skill, while omitting cross-skill
dependencies, intermediate states, and procedural constraints. Stage~2
therefore first reconstructs a richer compositional scenario, describes it
from five complementary dimensions, and then converts the resulting description
into aligned references:
\begin{equation}
    p_c
    \xrightarrow{\text{agentic reconstruction}}
    D_c
    \xrightarrow{\text{scenario synthesis}}
    C
    \xrightarrow{\text{reference building}}
    (R_S,R_I),
    \label{eq:stage2-flow}
\end{equation}
where $D_c$ is the five-dimensional scenario representation, $C$ is its
complete natural-language description, and $R_S$ and $R_I$ are the solution
and instruction references.

\paragraph{Agentic scenario reconstruction.}
We progressively reconstruct the scenario through five modules.
\emph{Skill analysis} identifies the capabilities, tools, inputs, outputs,
preconditions, and observable effects of each skill. \emph{Scenario
exploration} proposes concrete application settings in which the selected
skills can contribute to a shared user objective. \emph{Association and
filtering} retains scenarios that make meaningful use of the skills rather than
placing unrelated operations side by side. \emph{Evolution and recovery}
organizes the selected capabilities into a coherent workflow and recovers the
cross-skill dependencies, intermediate artifacts, and state transitions
connecting their operations. Finally, \emph{information expansion} enriches
the recovered workflow with concrete resources, formats, constraints, and
observable success conditions. Together, these modules transform a broad skill
combination into a compositional scenario with explicit interactions among its
capabilities.

\paragraph{Scenario representation and reference building.}
To preserve the reconstructed information, we separately describe the scenario
from five dimensions:
\begin{equation}
    D_c =
    \left\{
        d_{\mathrm{goal}},
        d_{\mathrm{context}},
        d_{\mathrm{capability}},
        d_{\mathrm{state}},
        d_{\mathrm{io\text{-}tool}}
    \right\}.
    \label{eq:scenario-dimensions}
\end{equation}
The \emph{goal} dimension defines the user objective and expected deliverables;
\emph{context} describes the application setting and motivation;
\emph{capability} specifies the role of each skill and its relationships with
other skills; \emph{state} records the initial, intermediate, and desired final
states; and \emph{inputs/outputs and tools} specifies the required files,
formats, schemas, paths, tools, services, and dependencies.

A model then integrates these five descriptions into a complete
natural-language scenario $C$. This integration preserves the information from
each dimension while expressing the task as a coherent user workflow. The
resulting description serves as the shared semantic reference for downstream
construction.

Based on $C$, we first generate the solution reference $R_S$, which records the
setup actions, execution workflow, intermediate artifacts, state transitions,
and key solution steps. We then generate the instruction reference $R_I$ from
both $C$ and $R_S$, specifying the task goal, inputs, outputs, deliverables, and
constraints:
\begin{equation}
    R_S=f_S(C), \qquad R_I=f_I(C,R_S).
    \label{eq:reference-building}
\end{equation}
A consistency-alignment model checks that the two references share the same
initial state and target outcome, that every instruction requirement is
supported by the solution workflow, and that every required effect is
observable in the final state. The complete scenario and its aligned references
are passed to Stage~3 as the shared specification for environment and task
construction.

Algorithm~\ref{alg:state-grounded-construction} summarizes Stage~3, from
environment realization to shared-state artifact generation and validation.

\begin{algorithm}[t]
\caption{Executable-state-grounded task construction}
\label{alg:state-grounded-construction}
\begin{algorithmic}[1]
\Require Reconstructed specification $Z=(C,R_S,R_I)$
\Ensure Validated task bundle $\mathcal{T}$ or failure

\State Plan an environment manifest from $Z$
\State Materialize the required files, services, dependencies, and assets
\State Retrieve and localize public resources when needed
\State Augment or perturb fixtures while preserving task semantics

\Repeat
    \State Build and initialize the environment
    \If{initialization fails}
        \State Repair the environment from the failure trace and $Z$
    \EndIf
\Until{the environment is valid or the repair budget is exhausted}

\If{the environment remains invalid}
    \State \Return failure
\EndIf

\State Observe the realized environment state $e_0$
\State Generate instruction $I$ from $R_I$ and $e_0$
\State Generate solution $S$ from $I$, $R_S$, and $e_0$
\State Execute $S$ to obtain the resulting state $e_T$
\State Generate verifier $V$ from $I$, $R_S$, $e_0$, and $e_T$
\State Package the task bundle $\mathcal{T}$

\Repeat
    \State Validate $\mathcal{T}$ from a clean initial state
    \If{validation fails}
        \State Identify the responsible artifact from the execution trace
        \State Repair only the identified artifact
    \EndIf
\Until{$\mathcal{T}$ is valid or the repair budget is exhausted}

\If{$\mathcal{T}$ is valid}
    \State \Return $\mathcal{T}$
\Else
    \State \Return failure
\EndIf

\end{algorithmic}
\end{algorithm}

\subsection{Executable-State-Grounded Task Construction}
\label{sec:state-grounded-construction}

Stage~3 converts the reconstructed specification
$Z=(C,R_S,R_I)$ into an executable task bundle. It first constructs and
repairs the execution environment, exposes the realized container state as
shared context for task-artifact generation, and finally validates and repairs
the complete task. Algorithm~\ref{alg:state-grounded-construction} summarizes
this procedure.

\paragraph{Environment construction and repair.}
Generating a complete environment in a single model response is unreliable for
file-rich tasks involving multiple documents, datasets, media assets, archives,
or binary files. We therefore separate environment planning from asset
materialization. Given $Z$, the environment agent first produces a manifest
describing the required directories, files, services, dependencies, and their
expected properties. It then materializes this manifest within a restricted
base image.

During materialization, the agent may use network access together with shell
and Python programs to retrieve public resources, transform them into the
required formats, or procedurally generate text and binary assets. Retrieved
resources are localized into the task build context so that the resulting
environment is self-contained and does not require network access during
evaluation. To avoid overly simple or template-like fixtures, the agent may
also augment and perturb collected or generated data while preserving the
intended schema and task semantics. For example, structured fixtures can be
expanded with additional records, metadata fields, distractor entries, and
cross-file relations. This produces sufficiently rich observable states for
constructing non-trivial tasks and verifiers.

The resulting environment contains the Dockerfile, fixture files, local
services, and dependencies. We build the image and execute initialization
checks before generating the final task artifacts. Compiler errors, missing
packages, malformed fixtures, failed downloads, and service failures are
returned to the environment agent for targeted repair. We allow at most three
environment-repair iterations, rebuilding and reinitializing the environment
after each repair. The repair process is conditioned on both the observed
failure trace and the reconstructed specification $Z$, preventing the agent
from removing task requirements merely to obtain a successful build.

\paragraph{Executable-state sharing.}
After the environment has been successfully built and initialized, we expose
its realized state as a shared grounding interface for all downstream artifact
generation. The instruction, solution, and verifier are generated sequentially,
with each generator receiving the reconstructed references and read access to
the same realized container state. The instruction is therefore grounded in
files, services, schemas, and resources that actually exist. The solution is
generated from the instruction and solution reference while directly inspecting
the same environment. The reference solution is subsequently executed to expose
the resulting final state, and the verifier is generated last using the
instruction, reference workflow, and observable initial and final states. We
favor behavioral and state-based checks over exact command matching so that
alternative correct solutions can pass.

The realized environment state serves as a shared coordination channel across
artifact generation. If environment construction or repair changes a filename,
path, port, package version, service configuration, input schema, or fixture
content, all downstream generators observe the updated state. This prevents the
instruction, solution, and verifier from being generated against different
implicit versions of the environment and reduces cross-artifact inconsistencies.

\paragraph{Validation and targeted repair.}
Each candidate is packaged in the Harbor format, containing
\texttt{environment/}, \texttt{solution/}, \texttt{tests/},
\texttt{instruction.md}, and \texttt{task.toml}. Validation checks that
(i) the environment builds and initializes successfully,
(ii) the verifier does not pass in the initial state,
(iii) the reference solution executes from a clean initial state, and
(iv) the verifier passes on the resulting final state.

When validation fails, a constrained router identifies the responsible
component from the execution trace and invokes only the corresponding repair
procedure. Build and initialization failures are assigned to the environment,
solution execution failures to the solution, test collection or assertion
defects to the verifier, and instruction--state mismatches to either the
instruction or environment according to the source of the inconsistency.
Targeted repair preserves valid components and avoids unnecessary regeneration
of the complete task bundle. Every repaired candidate is re-evaluated from a
clean initial state using the same validation procedure. We allow at most five
task-level repair iterations. Candidates that remain invalid after the repair
budget is exhausted are discarded. The complete construction funnel and repair
statistics are reported in Appendix~\ref{app:funnel}.

\begin{table}[t]
\caption{Trajectory- and task-level comparison of terminal-agent datasets.
Trajectories are collected using the Terminus-2 scaffold, and task performance
is evaluated using DeepSeek-V4-Pro with Terminus-2. Turns and Tests denote the
average interaction turns per trajectory and executable checkpoints per task.
Detailed protocols are provided in Appendix~\ref{app:dataset-comparison}.}
    \label{tab:dataset-comparison}
    \vspace{0.5em}
    \centering
    \small
    \renewcommand{\arraystretch}{0.95}
    \begin{tabular*}{\linewidth}
        {@{\extracolsep{\fill}}lcccccc@{}}
        \toprule
        \multirow{2}{*}{Dataset}
        & \multicolumn{2}{c}{Trajectory}
        & \multicolumn{4}{c}{Task} \\
        \cmidrule(lr){2-3}
        \cmidrule(l){4-7}
        & \#Traj.
        & Turns
        & \#Tasks
        & Tests
        & P@1
        & P@3 \\
        \midrule
        Nemotron-Terminal~\citep{pi2026dataengineering}
        & 5K
        & 6.12
        & 15K
        & 6.18
        & 40.67
        & 48.00 \\

        Endless-Terminals~\citep{gandhi2026endlessterminals}
        & 200
        & 4.53
        & 2,492
        & 5.51
        & 83.00
        & 87.00 \\

        Terminal-Lego~\citep{yang2026terminallego}
        & 32K
        & 5.77
        & 15K
        & 16.60
        & 47.00
        & 49.00 \\

        TerminalWorld~\citep{chu2026terminalworld}
        & 200
        & 11.94
        & 1,530
        & 3.98
        & 57.00
        & 82.00 \\

        Tmax~\citep{ivison2026tmax}
        & 500
        & 11.14
        & 15K
        & 3.29
        & 80.00
        & 86.00 \\

        \midrule
        \textbf{FACET (ours)}
        & 1.2K
        & 11.86
        & 6K
        & 22.77
        & 27.00
        & 35.00 \\
        \bottomrule
    \end{tabular*}
\end{table}
\section{Experiments}
\label{sec:experiments}

\subsection{Experimental Setup}
\label{sec:experimental-setup}

\paragraph{Models and training.}
We use the Terminus-2 agent, powered by DeepSeek-V4-Pro~\citep{deepseekai2026deepseekv4}, to generate rollouts on approximately 6K validated tasks. From these rollouts, we select 1.2K complete successful trajectories for supervised fine-tuning.
 We fine-tune
Qwen3.5-4B, Qwen3.5-9B, and Qwen3.5-27B~\citep{qwen2026qwen35} using
LLaMA-Factory~\citep{zheng2024llamafactory}.

\paragraph{Benchmark and evaluation protocol.}
We evaluate the base and fine-tuned models on Terminal-Bench 2.1, which uses
containerized tasks and execution-based verification
\citep{terminalbench2026v21}. All models use the Terminus-2 scaffold under the
same inference configuration. We run three attempts per task and report the
mean pass rate. Complete training and evaluation settings are provided in
Appendix~\ref{app:config}.

\subsection{Comparison with Existing Terminal Datasets}
\label{sec:dataset-comparison}

Table~\ref{tab:dataset-comparison} compares \benchname with existing
terminal-agent datasets from both trajectory- and task-level perspectives.
Detailed data sources, sampling procedures, and metric definitions are
provided in Appendix~\ref{app:dataset-comparison}.

\paragraph{Trajectory-level comparison.}
\benchname contains 1.2K training trajectories, fewer than
Nemotron-Terminal and Terminal-Lego, but its trajectories are comparatively
long. The average trajectory length of \benchname is 11.86 turns, close to
TerminalWorld at 11.94 turns and higher than Tmax (11.14),
Nemotron-Terminal (6.12), Terminal-Lego (5.77), and Endless-Terminals (4.53).
This result is consistent with our agentic scenario-reconstruction pipeline,
which combines related skills into workflows involving multiple dependent
operations. The resulting trajectories therefore capture relatively
long-horizon terminal interactions despite the smaller training-set size.

\paragraph{Task-level comparison.}
At the task level, \benchname contains 6,078 validated tasks and has the
largest number of executable tests per task, averaging 22.77. This exceeds
Terminal-Lego (16.60), Nemotron-Terminal (6.18), Endless-Terminals (5.51),
TerminalWorld (3.98), and Tmax (3.29). A larger number of executable tests
indicates that each task contains more independently checked requirements and
is evaluated against stricter completion criteria. This reflects our
artifact-aware construction pipeline, which verifies not only the primary
output but also secondary deliverables, content constraints, cross-artifact
consistency, and observable environment behavior.

Correspondingly, \benchname obtains 27.00 P@1 and 35.00 P@3, lower than the
results on the other datasets. The lower pass rates are consistent with the
larger number of task checkpoints: an agent must satisfy all required
conditions for the task to pass, and completing the main workflow alone is
insufficient if any checked requirement remains unmet. The increase from P@1
to P@3 shows that repeated attempts recover some failures, while the remaining
gap indicates that many tasks consistently expose requirement-satisfaction
errors.

\subsection{Main Results}
\label{sec:main-results}

Table~\ref{tab:results} reports the Terminal-Bench 2.1 performance of our
models together with representative systems reported in prior papers,
model reports, and our evaluations under the same setting.

\begin{table}[!t]
    \caption{Results on Terminal-Bench 2.1. Scores under our evaluation
    setting are averaged over three independent attempts per task.}
    \label{tab:results}
    \centering
    \small
    \setlength{\tabcolsep}{4pt}
    \renewcommand{\arraystretch}{0.96}

    \begin{tabular*}{\linewidth}{
        @{\extracolsep{\fill}}lccr@{}
    }
        \toprule
        Model & Size & Agent & Terminal-Bench 2.1 \\
        \midrule

        \multicolumn{4}{@{}l}{\emph{Reported reference models}} \\
        GPT-5.5 (xhigh)~\citep{terminalbench2026leaderboard}
            & --- & Terminus-2 & 78.00 \\
        Claude Opus 4.7 (max)~\citep{terminalbench2026leaderboard}
            & --- & Terminus-2 & 66.10 \\
        Gemini 3 Pro (high)~\citep{terminalbench2026leaderboard}
            & --- & Gemini CLI & 65.80 \\
        Intern-S2-Preview-397B~\citep{bai2026interns2}
            & 397B & Terminus-2 & 67.42 \\
        MiniMax M3~\citep{minimax2026m3}
            & 428B & Terminus-2 & 66.00 \\
        GLM-5.1 (max)~\citep{terminalbench2026leaderboard}
            & 744B & Claude Code & 58.70 \\

        \addlinespace[0.25em]
        \multicolumn{4}{@{}l}{\emph{Models evaluated under our setting}} \\
        Qwen3.6-27B
            & 27B & Terminus-2 & 53.93 \\
        Qwen3.5-397B-A17B
            & 397B & Terminus-2 & 49.06 \\
        Kimi-K2.6
            & 1T & Terminus-2 & 59.93 \\
        DeepSeek-V4-Pro-Preview (high)
            & 1.6T & Terminus-2 & 73.03 \\

        \addlinespace[0.25em]
        \multicolumn{4}{@{}l}{\emph{Qwen3.5 base models}} \\
        Qwen3.5-4B
            & 4B & Terminus-2 & 17.60 \\
        Qwen3.5-9B
            & 9B & Terminus-2 & 27.34 \\
        Qwen3.5-27B
            & 27B & Terminus-2 & 40.82 \\

        \midrule
        \multicolumn{4}{@{}l}{\emph{Fine-tuned models}} \\
        FACET-Terminal-Qwen3.5-4B
            & 4B
            & Terminus-2
            & \textbf{24.72}\,
              \textcolor{green!50!black}{(+7.12)} \\
        FACET-Terminal-Qwen3.5-9B
            & 9B
            & Terminus-2
            & \textbf{35.58}\,
              \textcolor{green!50!black}{(+8.24)} \\
        FACET-Terminal-Qwen3.5-27B
            & 27B
            & Terminus-2
            & \textbf{47.57}\,
              \textcolor{green!50!black}{(+6.75)} \\

        \bottomrule
    \end{tabular*}
\end{table}

Fine-tuning on only 1.2K successful trajectories yields consistent improvements
across all three model scales. Qwen3.5-4B improves from 17.60 to 24.72
(\textbf{+7.12}), Qwen3.5-9B from 27.34 to 35.58
(\textbf{+8.24}), and Qwen3.5-27B from 40.82 to 47.57
(\textbf{+6.75}). The 9B model achieves the largest absolute gain, while the
4B model shows the largest relative improvement of 40.5\%. These gains across
4B, 9B, and 27B models indicate that the synthesized supervision transfers
consistently across model scales rather than benefiting only a particular
capacity regime.

The improvement is also substantial when viewed across model scales. Our
27B model reaches 47.57 on Terminal-Bench 2.1, only 1.49 points below the
49.06 achieved by Qwen3.5-397B under the same evaluation setting, despite
using a model roughly 15$\times$ smaller. Moreover, fine-tuning the 27B model
closes most of the performance gap between its 40.82 base score and the much
larger Qwen3.5-397B. Together with the gains at 4B and 9B, these results
demonstrate that \benchname provides effective and data-efficient
supervision for improving terminal-agent capabilities.

\subsection{Task Analysis}
\label{sec:task-analysis}

\paragraph{Partial progress is common despite low end-to-end success.}
Terminal tasks impose conjunctive success criteria: an agent must satisfy all
required conditions for the task to pass, even when most of the workflow has
been completed correctly. This creates a substantial gap between local progress
and task-level success. Across teacher rollouts with parseable verifier results,
89.40\% of individual checks are satisfied, whereas only 20.94\% of completed
rollouts achieve full task success. We do not interpret check-level accuracy as
an alternative evaluation metric, since verifier checks differ in granularity
and importance. Rather, the gap indicates that unsuccessful trajectories often
contain substantial correct progress that is hidden by binary task rewards.

Figure~\ref{fig:failure-analysis-overview}(a) further reveals that these
failures are concentrated near the success boundary. Among unsuccessful
rollouts, 54.00\% fail only one or two verifier checks. Qualitative inspection
shows that such trajectories frequently construct the primary requested
artifact or complete the main workflow, but violate a small residual
requirement, such as an incorrect field value, an unmet constraint, or a
missing secondary deliverable. Thus, many failures reflect incomplete
requirement satisfaction rather than complete inability to solve the task.
Importantly, the number of failed checks should not be interpreted directly as
repair difficulty: a single failed assertion may encode a critical requirement,
while one underlying error may trigger several checks.

\paragraph{Difficulty emerges from compositional requirements.}
Task difficulty also varies with the structure of the requested workflow.
Across frequent skill categories, structured-data tasks tend to be solved more
reliably than narrative-document tasks, while performance generally decreases
for tasks with longer instructions and broader verifier coverage. These
patterns are consistent with the compositional nature of our tasks: longer
instructions typically introduce more atomic requirements, cross-file
dependencies, and output constraints, increasing the chance that an otherwise
successful trajectory misses at least one condition. Because these properties
are correlated, we treat them as descriptive characteristics rather than
isolated causal factors.

Taken together, the analysis suggests that the difficulty of our tasks lies
not only in executing the main workflow, but in satisfying a collection of
interdependent requirements precisely and completely. Dense executable
verification makes these residual errors observable, distinguishing
near-successful trajectories from failures that make little meaningful
progress. This fine-grained execution signal provides a richer characterization
of terminal-agent behavior than binary task outcomes alone. Detailed
task-level analyses are provided in
Appendix~\ref{app:task-outcome-analysis}.

\subsection{Analysis of generation schemes}
\label{sec:generation-scheme-analysis}

We compare three artifact-generation orders using the same 100
scenario--skill pairs. \emph{Forward}, adopted by our pipeline, generates the
instruction, solution, and solution-aware verifier sequentially
($I\rightarrow S\rightarrow V$) after constructing the environment.
\emph{Reverse} exchanges the order of the final two artifacts
($I\rightarrow V\rightarrow S$), generating the verifier from a shared textual
specification before the solution is available. \emph{Joint} produces all three
artifacts together in a single model call.

\begin{figure*}[t]
    \centering
    \begin{minipage}[c]{0.49\textwidth}
        \centering
        \includegraphics[width=0.98\linewidth]{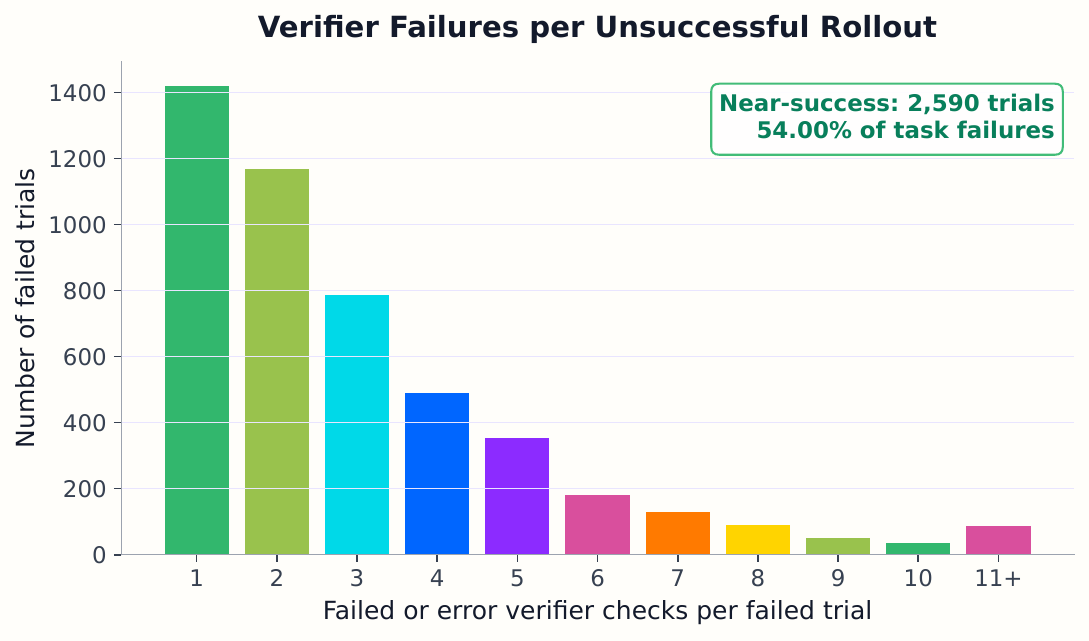}
        \vspace{-0.6em}
        \centerline{\small (a) Residual verifier failures.}
    \end{minipage}
    \hfill
    \begin{minipage}[c]{0.49\textwidth}
        \centering
        \includegraphics[width=0.98\linewidth]{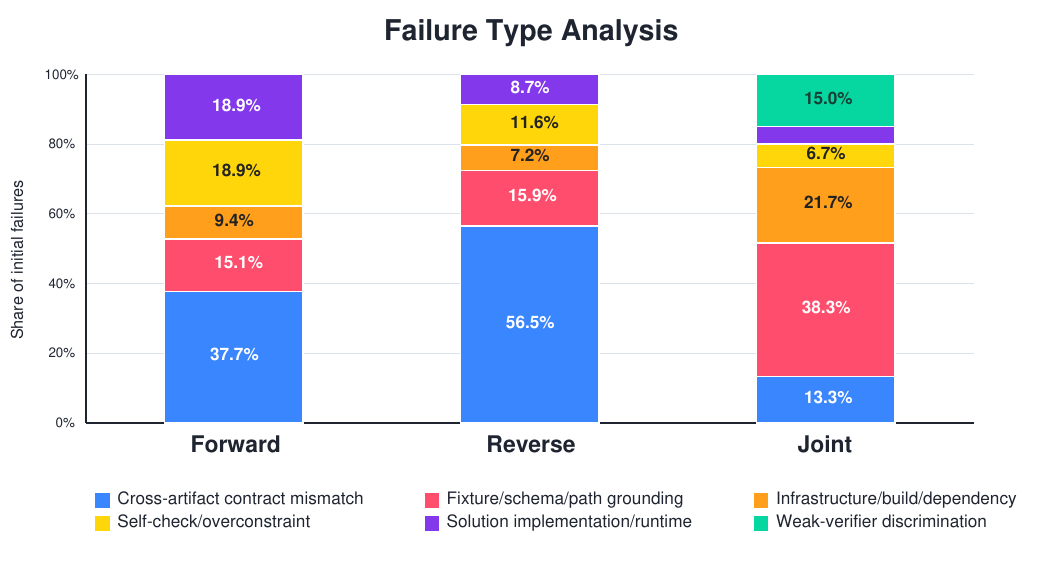}
        \vspace{-0.6em}
        \centerline{\small (b) Initial failure composition by generation scheme.}
    \end{minipage}
    \caption{Analysis of execution and synthesis failures.
    (a) Distribution of failed or errored verifier checks among unsuccessful
    teacher rollouts; three rollouts without a parsed
    \texttt{FAILED}/\texttt{ERROR} result are omitted.
    (b) Distribution of initial validation failure types under the Forward,
    Reverse, and Joint generation schemes.}
    \label{fig:failure-analysis-overview}
\end{figure*}

Forward, Reverse, and Joint deliver 99, 91, and 96 tasks to validation,
respectively, of which 46, 22, and 36 are initially valid. Their corresponding
initial validity rates are 46.5\%, 24.2\%, and 37.5\%.
Figure~\ref{fig:failure-analysis-overview}(b) further shows that generation
order changes not only the overall validity but also where synthesis failures
occur. Cross-artifact contract mismatch accounts for 56.5\% of Reverse
failures, suggesting that generating the verifier before observing the solution
makes behavioral alignment more difficult. Joint reduces contract mismatch to
13.3\%, but shifts errors toward fixture/schema/path grounding (38.3\%) and
infrastructure or dependency failures (21.7\%). Forward achieves the highest
initial validity while exhibiting a more distributed failure profile, with
contract mismatch accounting for 37.7\%.

The advantage persists after repair. Under the recorded repair settings,
Forward recovers 37 of its 53 initial failures and reaches a final yield of
83/100 tasks. Reverse recovers 41 of 69 failures and reaches 63/100, while
Joint recovers 29 of 60 and reaches 65/100. Since Forward permits five repair
rounds whereas Reverse and Joint permit three, these final yields characterize
the complete pipeline configurations rather than repair efficiency under an
identical budget.

A paired comparison over the 88 scenario--skill pairs that reach validation
under all three schemes provides a more controlled comparison. Forward succeeds
on 29 pairs where Reverse fails, compared with only 9 pairs in the opposite
direction ($p=0.0017$, two-sided exact sign test). The corresponding
Forward--Joint counts are 27 and 18 ($p=0.233$). These results provide strong
evidence that generating the solution before the verifier improves
cross-artifact alignment over the Reverse order, while the difference between
Forward and Joint is less conclusive. Overall, the results support the
sequential, environment-grounded generation order used in \benchname,
particularly for maintaining alignment between the generated solution and its
executable verifier. Appendix~\ref{app:failure-analysis} reports the complete
metric definitions, coverage statistics, and evaluation qualifications.

\section{Related Work}
\label{sec:related}

\paragraph{Terminal benchmarks and environments.}
Terminal-Bench 2.0 and 2.1 evaluate agents on realistic, containerized
command-line tasks with execution-based verification
\citep{merrill2026terminalbench,terminalbench2026v21}, while TerminalWorld
constructs validated tasks from real terminal recordings
\citep{chu2026terminalworld}. Harbor provides a common format for packaging
tasks, executing agents, and collecting rollouts
\citep{harbor2026software}; our generated tasks follow this format directly.

\paragraph{Scalable task synthesis.}
Existing pipelines construct terminal tasks through procedural generation,
domain specifications, seed datasets, skill composition, and
requirement-driven retrieval
\citep{gandhi2026endlessterminals,peng2026litecoderterminal,
pi2026dataengineering,ivison2026tmax,zhao2026nexforge}.
Agent Skills provide portable procedural knowledge
\citep{agentskills2025specification}; SkillSynth organizes skills into
scenario-mediated graphs \citep{fan2026skillsynth}; and Terminal-Lego studies
environment-grounded trajectory quality \citep{yang2026terminallego}.
Building on these directions, our work focuses on reconstructing complex
scenarios from heterogeneous skills and coordinating task artifacts through a
shared, realized environment state.

\section{Conclusion}

We presented \benchname, a framework for synthesizing complex and verifiable
terminal tasks from heterogeneous agent skills. \benchname reconstructs related
skills into coherent, information-rich scenarios, realizes and repairs the
execution environment before finalizing task artifacts, and grounds the
instruction, solution, and verifier in the same executable state. Together
with targeted validation and repair, these designs improve consistency across
task components while preserving the information and constraints inherited
from the source skills.

Our experiments demonstrate the effectiveness of both the synthesis pipeline
and the resulting supervision. The analysis of generation schemes shows that
sequential, environment-grounded construction achieves the highest task yield
and improves solution--verifier alignment compared with generating the verifier
before the solution. More importantly, supervised fine-tuning on the resulting
successful trajectories consistently improves Qwen3.5 models from 4B to 27B
on Terminal-Bench 2.1, with absolute gains of 7.12, 8.24, and 6.75 points.
The resulting 27B model reaches 47.57, approaching the 49.06 performance of
the substantially larger Qwen3.5-397B under the same evaluation setting.

These results suggest that carefully coordinating scenario reconstruction,
executable-state grounding, and artifact-level validation can provide
data-efficient supervision for terminal agents without relying solely on
large-scale task generation. Looking forward, we plan to extend \benchname to
broader sources of procedural knowledge and more diverse interactive
environments, and to investigate how the generated tasks and execution
feedback can support reinforcement learning and continued agent improvement.

\newpage
\subsection*{AI Use Statement}

Generative AI tools were used to assist with drafting, language editing, and
LaTeX preparation. The authors are responsible for checking all source
attributions, experimental records, numerical claims, and generated text, and
take responsibility for the final content of the paper.

\bibliography{iclr2027_conference}
\bibliographystyle{iclr2027_conference}

\newpage
\appendix

\section{Dataset Details}
\label{app:data}

\subsection{Source-skill taxonomy}

We organize the 71,341 retained skills into five top-level families and 34
fine-grained categories. Table~\ref{tab:skills} reports the category counts and
proportions, computed over the complete corpus without sampling. We separately
classify the 6,078 validated tasks into nine task families to characterize the
coverage of the synthesized dataset.

\begin{table}[h]
    \caption{Top-level source-skill distribution.}
    \label{tab:skills}
    \centering
    \small
    \begin{tabular}{lrr}
        \toprule
        Category & Skills & Share \\
        \midrule
        AI, agents, and tools & 15,182 & 21.28\% \\
        Software, systems, and security & 15,059 & 21.11\% \\
        Data, analysis, and research & 12,409 & 17.39\% \\
        Documents, productivity, and workflows & 11,267 & 15.79\% \\
        Multimedia, creation, and publishing & 17,424 & 24.42\% \\
        \midrule
        Total & 71,341 & 100.00\% \\
        \bottomrule
    \end{tabular}
\end{table}
\begin{table*}[t]
    \caption{Fine-grained source-skill categories. ``Global'' is the share of
    all 71,341 skills; ``within parent'' is the share inside the corresponding
    top-level category.}
    \label{tab:skilldet}
    \centering
    \scriptsize
    \begin{tabular}{p{0.22\textwidth}p{0.32\textwidth}rrr}
        \toprule
        Parent & Fine-grained category & Count & Global & Within parent \\
        \midrule
        AI, agents, and tools & Agent orchestration and automation & 2,782 & 3.90\% & 18.32\% \\
        & Prompt and model calls & 3,310 & 4.64\% & 21.80\% \\
        & Skills, plugins, and extensions & 1,637 & 2.29\% & 10.78\% \\
        & MCP and external tools & 2,579 & 3.62\% & 16.99\% \\
        & Memory, RAG, and knowledge bases & 3,439 & 4.82\% & 22.65\% \\
        & Multi-agent collaboration & 1,435 & 2.01\% & 9.45\% \\
        \midrule
        Software, systems, and security & Code generation and development & 3,065 & 4.30\% & 20.35\% \\
        & Testing, debugging, and code quality & 2,167 & 3.04\% & 14.39\% \\
        & Git, build, and dependency management & 1,904 & 2.67\% & 12.64\% \\
        & Deployment, containers, and DevOps & 2,069 & 2.90\% & 13.74\% \\
        & System administration and CLI & 3,576 & 5.01\% & 23.75\% \\
        & Security, privacy, and compliance & 2,278 & 3.19\% & 15.13\% \\
        \midrule
        Data, analysis, and research & JSON, YAML, and XML & 2,752 & 3.86\% & 22.18\% \\
        & CSV, Excel, and spreadsheets & 1,544 & 2.16\% & 12.44\% \\
        & Databases and SQL & 1,627 & 2.28\% & 13.11\% \\
        & Data cleaning, conversion, and validation & 1,595 & 2.24\% & 12.85\% \\
        & Statistical analysis and metrics & 1,892 & 2.65\% & 15.25\% \\
        & Visualization and dashboards & 1,419 & 1.99\% & 11.44\% \\
        & Search, research, and extraction & 1,580 & 2.21\% & 12.73\% \\
        \midrule
        Documents, productivity, and workflows & Documents and Markdown & 1,496 & 2.10\% & 13.28\% \\
        & Reports, summaries, and briefs & 1,865 & 2.61\% & 16.55\% \\
        & PDF, Office, and presentations & 1,429 & 2.00\% & 12.68\% \\
        & Office and personal productivity & 1,592 & 2.23\% & 14.13\% \\
        & Project, task, and schedule management & 1,682 & 2.36\% & 14.93\% \\
        & System integration and automation & 1,597 & 2.24\% & 14.17\% \\
        & Audit, checklists, and operation records & 1,606 & 2.25\% & 14.25\% \\
        \midrule
        Multimedia, creation, and publishing & Image generation and editing & 2,699 & 3.78\% & 15.49\% \\
        & Design, drawing, and visual assets & 1,858 & 2.60\% & 10.66\% \\
        & Audio, speech, and music & 1,917 & 2.69\% & 11.00\% \\
        & Video, animation, and captions & 2,491 & 3.49\% & 14.30\% \\
        & Content writing and creative generation & 1,731 & 2.43\% & 9.93\% \\
        & Social media and community operations & 1,479 & 2.07\% & 8.49\% \\
        & SEO, websites, and content publishing & 3,778 & 5.30\% & 21.68\% \\
        & Marketing campaigns and channels & 1,471 & 2.06\% & 8.44\% \\
        \bottomrule
    \end{tabular}
\end{table*}

\begin{figure*}[t]
    \centering
    \begin{minipage}[t]{0.49\textwidth}
        \centering
        \includegraphics[width=0.98\linewidth]{
            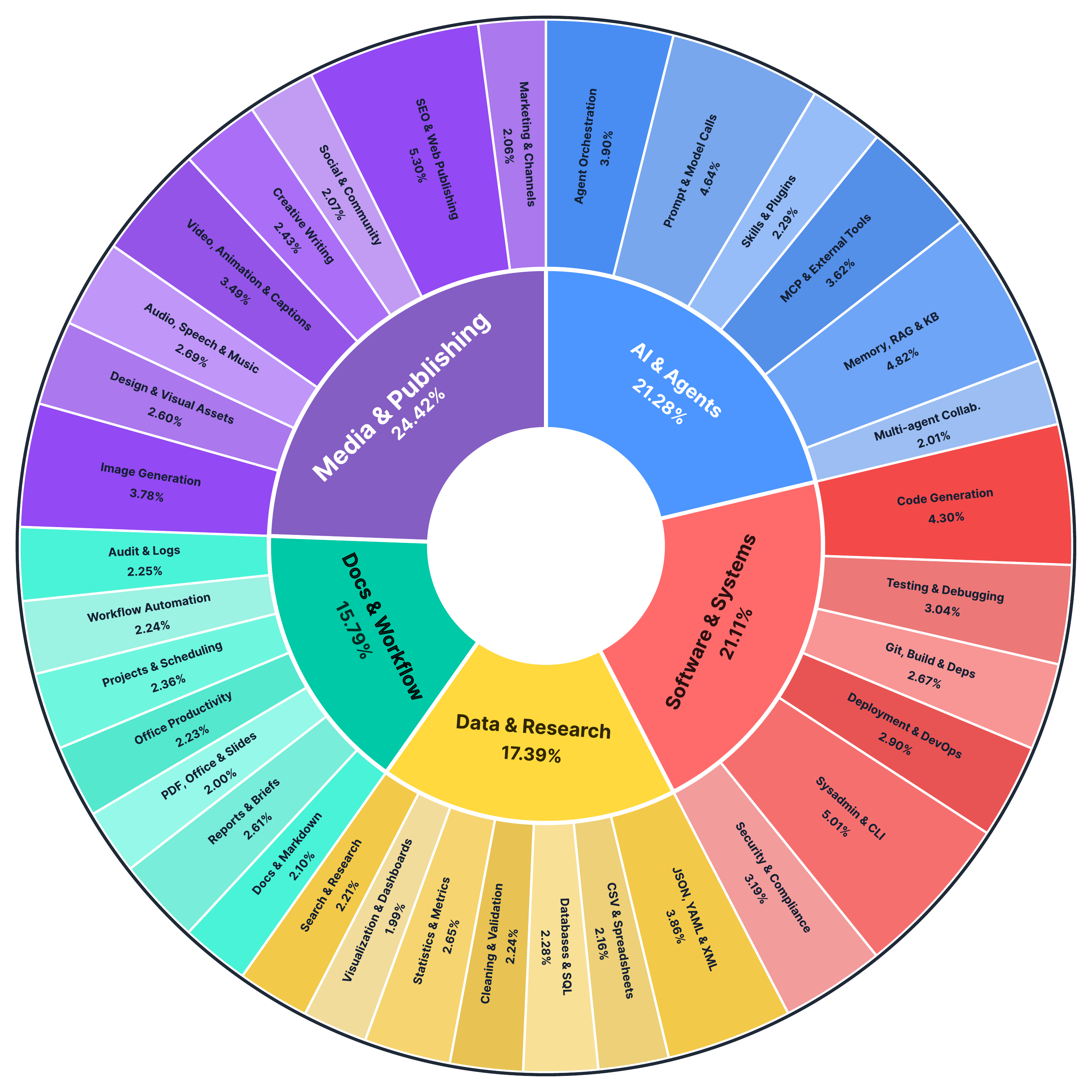
        }
        \vspace{-0.8em}
        \centerline{\small (a) Source-skill distribution.}
    \end{minipage}
    \hfill
    \begin{minipage}[t]{0.49\textwidth}
        \centering
        \includegraphics[width=0.98\linewidth]{
            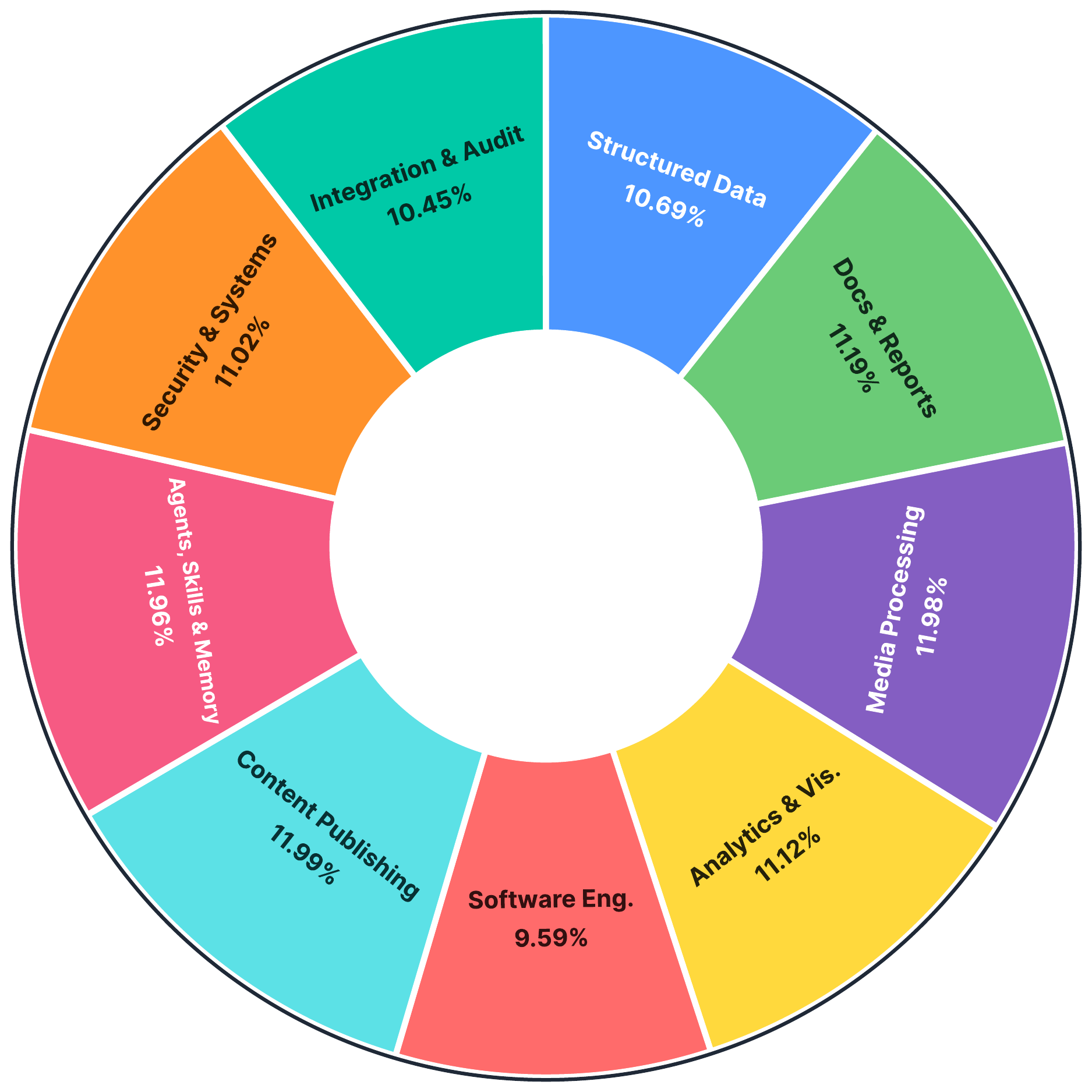
        }
        \vspace{-0.8em}
        \centerline{\small (b) Synthesized-task distribution.}
    \end{minipage}
    \caption{Distributions of source skills and synthesized tasks.
    (a) The retained skill corpus spans five top-level families and 34
    fine-grained categories. (b) The 6,078 validated tasks are distributed
    across nine task families, whose individual shares range from 9.59\% to
    11.99\%.}
    \label{fig:distributions}
\end{figure*}

\subsection{Task and rollout accounting}

The final dataset contains 6,078 tasks, of which 6,074 have corresponding
rollout records; four jobs fail to produce a record. Among the recorded runs,
1,270 receive reward 1, 4,796 complete execution but fail at least one task
requirement, and eight terminate because of infrastructure errors. Excluding
infrastructure failures yields 6,066 completed runs and a teacher success rate
of $1,270/6,066=20.94\%$.

Of the completed runs, 6,064 contain parseable pytest collection summaries and
are included in the check-level analysis in
Section~\ref{sec:task-analysis}. Failed trajectories are retained for error
analysis but excluded from supervised fine-tuning. From the successful
rollouts, we select 1,200 complete trajectories for the final SFT dataset.

\subsection{Task outcomes and skill-tag variation}
\label{app:task-outcome-analysis}

Section~\ref{sec:task-analysis} reports the aggregate task- and check-level
results. Here, we examine how task-level success varies across skill tags and
broader task characteristics.

\begin{figure*}[t]
    \centering
    \includegraphics[width=0.88\textwidth]{
        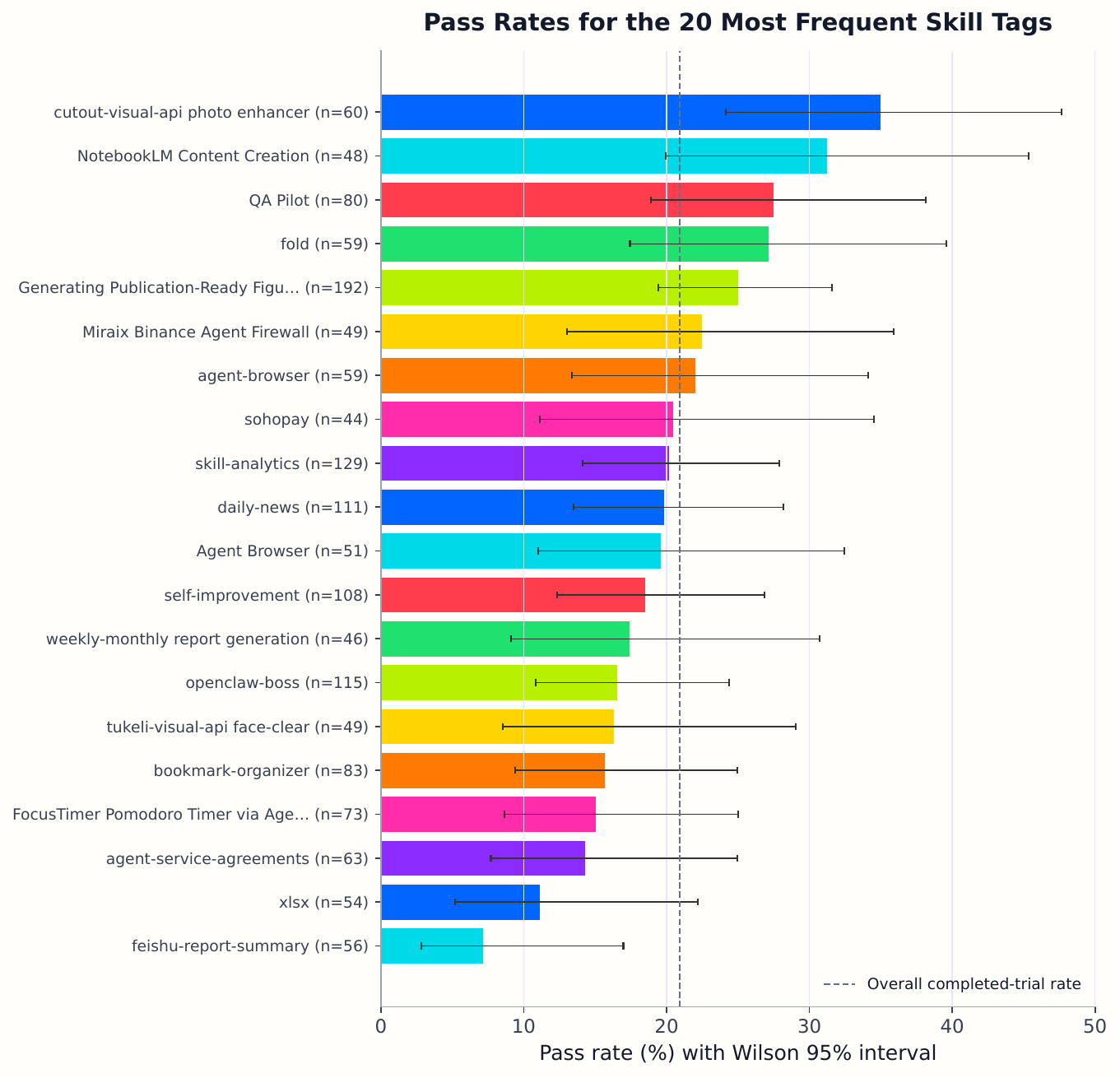
    }
    \caption{Strict task-level pass rates for the 20 most frequent skill tags
    among the 6,066 completed rollouts. Error bars indicate Wilson 95\%
    confidence intervals, and the dashed line denotes the overall pass rate.
    Tags are selected by frequency and ordered by observed pass rate.}
    \label{fig:skill-tag-pass-rates}
\end{figure*}

Figure~\ref{fig:skill-tag-pass-rates} shows considerable variation across
frequent skill tags, with observed pass rates ranging from 7.14\% to 35.00\%.
Several estimates have wide confidence intervals because their sample sizes
are modest. Each task is associated with seven skill tags, so the groups
overlap and their estimates are not statistically independent. Moreover, tags
co-vary with task topic, output type, instruction length, and other tags.
These results should therefore be interpreted as descriptive indicators of
task difficulty rather than causal estimates for individual skills.

We additionally group tasks by output structure, instruction length, and
verifier breadth. Structured-data tasks have higher observed pass rates than
narrative-document tasks, while success generally decreases for longer
instructions and broader verifier suites. These properties are correlated:
longer instructions typically introduce more atomic requirements, output
constraints, and cross-file dependencies, whereas broader verifiers increase
the likelihood that a single omission causes the task to fail. The observed
trends therefore do not isolate a single source of difficulty, but consistently
identify requirement tracking and final-state consistency as important
challenges in complex terminal tasks.

\subsection{Construction funnel}
\label{app:funnel}

Table~\ref{tab:funnel} summarizes the number of candidate tasks retained
through environment construction and task validation. Before validation, 58
candidates are excluded for reasons including weak verifiers, ambiguous
deliverables, unresolved external dependencies, and workflows that permit
trivial bypasses. Since these exclusion criteria may overlap, we report their
aggregate count rather than a mutually exclusive per-category breakdown.

\begin{table}[t]
    \caption{Task-construction funnel. Percentages in the last column use the
    immediately preceding comparable stage.}
    \label{tab:funnel}
    \centering
    \small
    \begin{tabular}{lrr}
        \toprule
        Stage & Count & Stage retention \\
        \midrule
        Scenario--skill seeds & 7,852 & --- \\
        Seeds with first-build logs & 7,841 & 99.86\% \\
        Initial environment success & 6,630 & 84.56\% \\
        Environment repair recovery & 874 & --- \\
        Successful environments & 7,504 & 95.70\% \\
        Entering task validation & 7,446 & 99.23\% \\
        First-pass valid tasks & 2,856 & 38.35\% \\
        Task repair recovery & 3,222 & --- \\
        Final validated tasks & 6,078 & 81.63\% \\
        \bottomrule
    \end{tabular}
\end{table}

\subsection{Dataset-comparison details}
\label{app:dataset-comparison}

Table~\ref{tab:dataset-comparison} reports trajectory statistics,
task-level verifier statistics, and common-solver evaluation results for
existing terminal-agent datasets
\citep{pi2026dataengineering,gandhi2026endlessterminals,
yang2026terminallego,ivison2026tmax,chu2026terminalworld}.
Because these quantities are computed from different underlying units, we
describe their data sources and computation procedures separately.

\subsection{Dataset-comparison details}
\label{app:dataset-comparison}

Table~\ref{tab:dataset-comparison} reports trajectory-level statistics,
task-level statistics, and common-solver evaluation results for existing
terminal-agent datasets
\citep{pi2026dataengineering,gandhi2026endlessterminals,
yang2026terminallego,ivison2026tmax,chu2026terminalworld}.
The three groups of columns describe related but distinct aspects of each
dataset, and are therefore computed from different units of analysis.

\paragraph{Relationship between trajectory- and task-level data.}
A task is an executable problem instance containing an instruction, an initial
environment, and an executable verifier. A trajectory is an agent interaction
record produced while attempting one such task. The two units are related
because every trajectory is generated from a task, but they are not
necessarily paired one-to-one: a task may have multiple rollout attempts, no
available trajectory, or a trajectory excluded by filtering, while a released
trajectory collection may cover only a subset of the corresponding task set.

Consequently, the Trajectory columns in
Table~\ref{tab:dataset-comparison} characterize the available or recollected
agent interactions, whereas the Task columns characterize the executable task
collection itself. The trajectory sample used to compute average turns, the
task sample used to compute average tests, and the 100-task sample used to
compute P@1 and P@3 are selected independently. Statistics across these column
groups should therefore be interpreted as dataset-level characteristics
rather than measurements over exactly the same task instances. To improve
comparability, all recollected trajectories and common-solver evaluations use
the same Terminus-2 scaffold, and random sampling uses seed 42.

\paragraph{Trajectory data and metrics.}
For Nemotron-Terminal and Terminal-Lego, we use their released Terminus-2
trajectory collections, containing approximately 5K and 32K trajectories,
respectively. For Endless-Terminals and TerminalWorld, we randomly sample 200
tasks from each dataset and collect one rollout per task. For Tmax, we sample
500 tasks and recollect rollouts with Terminus-2 instead of using its released
mini-SWE-agent trajectories. Recollected rollouts are generated with
DeepSeek-V4-Pro, while the \benchname statistics are computed from the 1,200
complete successful trajectories used for supervised fine-tuning.

The \#Traj.\ column reports the number of trajectories included in each
sample. A turn corresponds to one assistant interaction step and its resulting
environment response. Turns is the average number of turns across successful,
parseable trajectories; infrastructure failures, missing records, and
unparseable trajectories are excluded.

\paragraph{Task data and verifier metrics.}
Task-level statistics are computed from executable task bundles independently
of rollout success. We use up to 15,000 tasks per dataset, sampling with seed
42 when necessary. This gives 15,000-task samples for Nemotron-Terminal,
Terminal-Lego, and Tmax, together with the complete collections of
Endless-Terminals (2,492), TerminalWorld (1,530), and \benchname (6,078).

The \#Tasks column reports the number of tasks included in the analysis. For
each task, we execute its verifier through the Harbor testing procedure and
record the number of test items collected by \texttt{pytest}. Tests is
calculated by summing these per-task counts and dividing by the number of
analyzed tasks. It therefore represents the average number of executable
checkpoints per task. A higher value indicates that more aspects of task
completion are checked and that the task is evaluated against stricter
requirements. We count collected \texttt{pytest} items rather than individual
assertions, since one test item may contain multiple assertions.

\paragraph{Common-solver evaluation metrics.}
P@1 and P@3 are computed from 100 tasks sampled from each dataset using seed
42. DeepSeek-V4-Pro attempts every task three times with Terminus-2, and each
attempt starts from a clean environment. P@1 is the success rate across all
300 individual attempts, while P@3 is the percentage of tasks solved in at
least one of the three attempts.

\paragraph{Comparison scope.}
The shared scaffold, solver, attempt count, and sampling procedure reduce
evaluation-side differences. The results nevertheless remain descriptive
dataset-level comparisons because the datasets differ in their domains,
construction procedures, task distributions, and verifier designs.

\subsection{Command-level behavior in successful trajectories}
\label{app:trajectory-behavior}

We analyze the 1,270 parseable teacher trajectories that receive reward 1,
comprising 15,075 assistant turns and 39,136 shell-command occurrences.
Commands are classified using both their names and execution contexts. Reads,
searches, listings, and checks are treated as observations, whereas writes,
installation, movement, deletion, and build operations are treated as
actions. For multipurpose commands such as \texttt{cat} and
\texttt{python3}, the classification additionally considers arguments,
redirection, and script content.

\begin{figure*}[t]
    \centering
    \begin{minipage}[t]{0.49\textwidth}
        \centering
        \includegraphics[
            height=0.29\textheight,
            width=0.98\linewidth,
            keepaspectratio
        ]{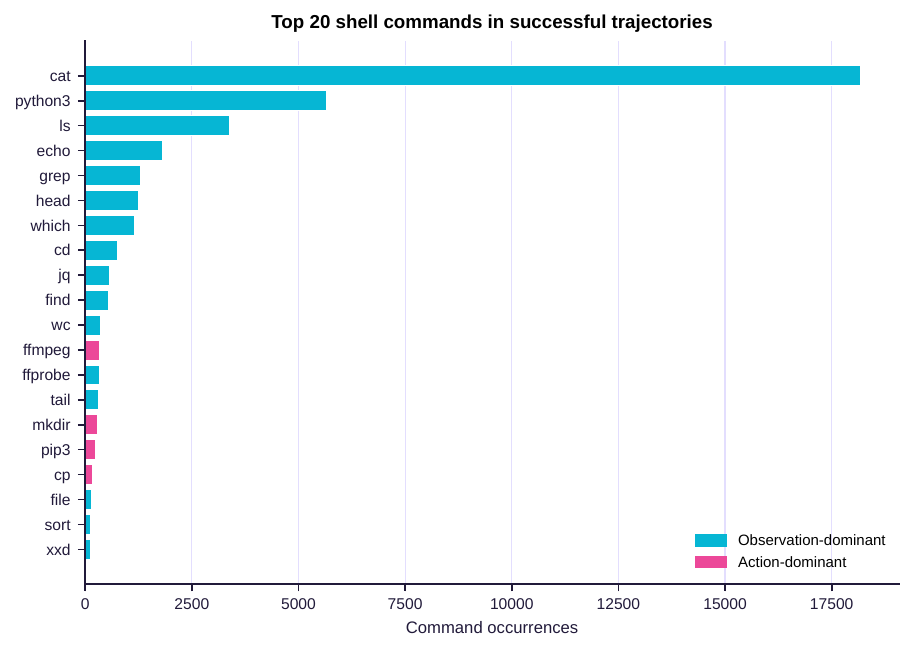}
        \vspace{-0.6em}
        \centerline{\small (a) Command-occurrence frequency.}
    \end{minipage}
    \hfill
    \begin{minipage}[t]{0.49\textwidth}
        \centering
        \includegraphics[
            height=0.29\textheight,
            width=0.98\linewidth,
            keepaspectratio
        ]{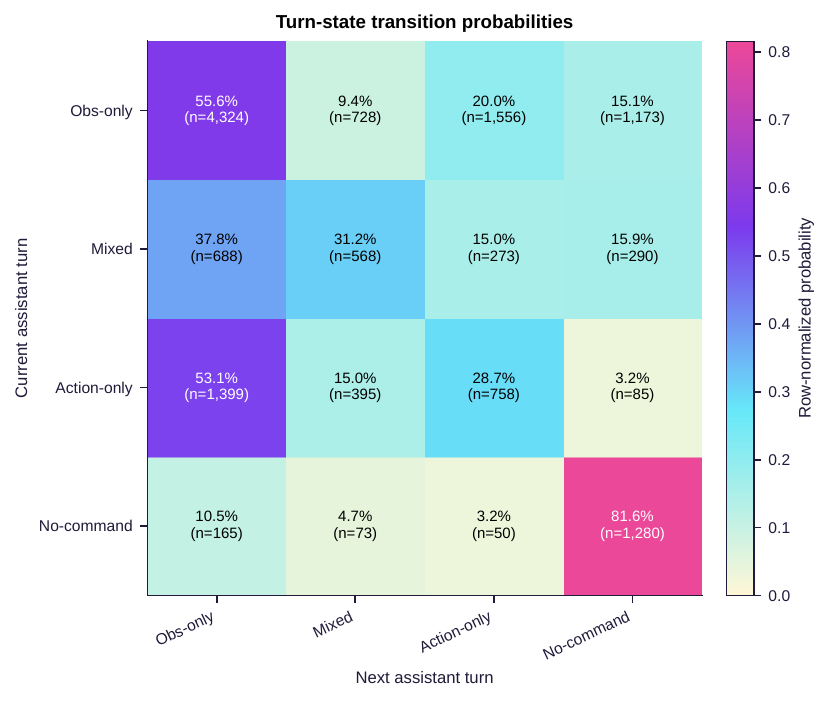}
        \vspace{-0.6em}
        \centerline{\small (b) Adjacent-turn transition probabilities.}
    \end{minipage}
    \caption{Command-level patterns in successful teacher trajectories.
    The left panel shows the most frequent shell commands, colored by their
    dominant contextual class. The right panel reports row-normalized
    transitions between adjacent assistant-turn states.}
    \label{fig:trajectory-behavior}
\end{figure*}

Figure~\ref{fig:trajectory-behavior}(a) shows that successful trajectories use
a concentrated terminal vocabulary. The commands \texttt{cat},
\texttt{python3}, and \texttt{ls} account for 69.5\% of all command
occurrences, while the ten most frequent commands account for 88.4\%. This
pattern suggests that the teacher repeatedly uses a compact interaction
interface---file inspection, script-based artifact construction, and output
verification---across otherwise heterogeneous tasks. Command names alone do
not uniquely determine behavior: for example, \texttt{cat} may either inspect
a file or create one through redirection.

Figure~\ref{fig:trajectory-behavior}(b) provides complementary turn-level
evidence of an observe--act--verify loop. Most successful trajectories begin
with an observation-only turn. Following an action-only turn, the next turn
returns to observation in 53.1\% of transitions, compared with 28.7\% that
continue directly to another action-only turn. Observation-only turns also
frequently occur consecutively, indicating that the agent often performs
multiple checks before modifying the environment.

Table~\ref{tab:trajectory-behavior} summarizes the command- and turn-level
statistics supporting these observations.

\begin{table}[t]
    \caption{Selected command- and turn-level statistics for successful teacher
    trajectories. Shares for command statistics use all 39,136 command
    occurrences; transition probabilities are normalized within the current
    turn state.}
    \label{tab:trajectory-behavior}
    \centering
    \small
    \begin{tabular}{llr}
        \toprule
        View & Statistic & Value \\
        \midrule
        \multirow{4}{*}{Command usage}
        & \texttt{cat} occurrences & 18,168 (46.4\%) \\
        & Top three commands & 27,207 (69.5\%) \\
        & Top ten commands & 34,598 (88.4\%) \\
        & Observation commands & 33,140 (84.7\%) \\
        \midrule
        \multirow{4}{*}{Turn dynamics}
        & Observation-only first turn & 1,214 (95.6\%) \\
        & Observation-only $\rightarrow$ observation-only & 4,324 (55.6\%) \\
        & Action-only $\rightarrow$ observation-only & 1,399 (53.1\%) \\
        & Action-only $\rightarrow$ action-only & 758 (28.7\%) \\
        \bottomrule
    \end{tabular}
\end{table}

These results characterize successful trajectories but do not establish which
behaviors cause success. The command parser cannot fully recover dynamically
generated shell operations, and comparison with failed trajectories would be
required to determine whether the observed patterns reliably distinguish
successful from unsuccessful behavior.

\section{Generation-Scheme Failure Analysis}
\label{app:failure-analysis}

The three generation schemes begin from the same 100 semantic path IDs. To
remain consistent with Figure~\ref{fig:failure-analysis-overview}(b) and the
associated result table, we retain the original scheme names: Base denotes
\emph{staged generation}, Hint-bundle v2 denotes \emph{contract-first
generation}, and Single-bundle v3 denotes \emph{joint generation}.

Initial validity requires a task to pass the complete validation sequence
before repair. Oracle feasibility requires the reference solution to execute
successfully from a clean initial state. Negative discrimination additionally
requires incomplete or partial solutions to fail the verifier. Final yield
counts all tasks recovered within the repair budget assigned to each scheme.

\begin{table}[t]
    \caption{Outcomes of three artifact-generation orders on 100 shared
    semantic paths. Initial validity is computed over tasks that reach
    validation, while final yield is computed over all selected paths.}
    \label{tab:abl}
    \centering
    \small
    \begin{tabular}{lccc}
        \toprule
        Scheme & Reached validation & Initially valid & Final yield \\
        \midrule
        Forward (Ours) & 99 & 46 (46.5\%) & 83/100 \\
        Reverse        & 91 & 22 (24.2\%) & 63/100 \\
        Joint          & 96 & 36 (37.5\%) & 65/100 \\
        \bottomrule
    \end{tabular}
\end{table}

The schemes operate under different repair budgets. Staged generation permits
five repair rounds, whereas contract-first and joint generation permit three.
Staged generation repairs 37 of its 53 initial failures and reaches a final
yield of 83/100. Contract-first generation repairs 41 of 69 and reaches
63/100, while joint generation repairs 29 of 60 and reaches 65/100. These
values describe the end-to-end output of each recorded pipeline configuration;
they should not be interpreted as a controlled comparison of repair efficiency
per iteration.

For the paired comparison, we restrict the analysis to the 88 semantic paths
that reach validation under all three schemes. Staged generation succeeds
alone against contract-first generation on 29 paths, while contract-first
generation succeeds alone on 9 paths. A two-sided exact sign test gives
$p=0.0017$. Staged generation succeeds alone against joint generation on 27
paths, while joint generation succeeds alone on 18, giving $p=0.233$.

Staged and joint generation have similar observed oracle feasibility:
46 of 99 staged tasks and 45 of 96 joint tasks have executable reference
solutions before repair. Joint generation additionally exposes nine
weak-verifier cases in which an incomplete solution passes. However,
partial-solution coverage is uneven across schemes: 95/96 for joint generation,
7/99 for staged generation, and 9/91 for contract-first generation.
Negative-discrimination results are therefore reported descriptively and
should not be directly compared without standardized partial-solution
coverage.

The generation-cost analysis measures only the artifact-generation branches.
Joint generation uses one model call with 2.80 minutes of referenced call
latency per task. Staged and contract-first generation use three calls, with
average latencies of 4.14 and 5.19 minutes, respectively. These measurements
exclude shared-prefix processing, validation, and repair, and therefore serve
as generation-cost proxies rather than complete end-to-end runtime estimates.

\section{Task Construction and Validation Details}
\label{app:format}

\subsection{Harbor task layout}

Each generated task follows the Harbor directory structure:

\begin{verbatim}
task/
  instruction.md       # user-visible request
  task.toml             # runtime metadata
  environment/          # Dockerfile and fixtures
  solution/             # executable reference solution
  tests/                # verifier and test assets
\end{verbatim}

\subsection{Shared-state construction and Docker validation}

The environment is built and repaired before the final task artifacts are
generated. After the image starts successfully, the system inspects the task
workspace and records the realized initial state $e_0$, including available
files, directories, schemas, dependencies, and local services. The instruction,
solution, and verifier generators receive the same read-only state record,
ensuring that they refer to a consistent environment without sharing a mutable
container.

Each completed task then undergoes the Docker round-trip procedure summarized
in Table~\ref{tab:docker-validation}. Baseline and oracle validation are
performed in independent clean containers to prevent state leakage between
trials.

\begin{table*}[t]
    \caption{Shared-state task construction and Docker round-trip validation.}
    \label{tab:docker-validation}
    \centering
    \small
    \begin{tabular}{p{0.06\textwidth}p{0.20\textwidth}p{0.66\textwidth}}
        \toprule
        Step & Stage & Operation \\
        \midrule
        1 & Materialize environment &
        Generate the Dockerfile, fixtures, dependencies, and initialization
        scripts from the reconstructed task specification. \\

        2 & Build and repair &
        Build the image with \texttt{docker build}. Build or initialization
        failures are returned to the environment-repair agent for at most
        three iterations. \\

        3 & Capture shared state &
        Start a temporary container, inspect the task workspace, and record
        the realized initial state $e_0$. \\

        4 & Generate artifacts &
        Generate the final instruction, reference solution, and verifier using
        the same reconstructed specification and shared state $e_0$. \\

        5 & Baseline validation &
        Start a clean container, copy and execute the verifier without running
        the solution, and require reward 0. \\

        6 & Oracle validation &
        Start another clean container, copy and execute
        \texttt{solution/solve.sh}, run \texttt{tests/test.sh}, and require
        reward 1. \\

        7 & Repair and revalidate &
        Classify a failure as an instruction, environment, solution, or
        verifier defect, apply the corresponding repair, and repeat the full
        validation procedure for at most five rounds. \\

        8 & Accept and clean up &
        Retain the task only after all validation conditions pass, then stop
        and remove temporary containers and images. \\
        \bottomrule
    \end{tabular}
\end{table*}

A task is accepted only if its image builds successfully, its verifier rejects
the untouched initial state, and its reference solution passes from a clean
environment. Every repair triggers the complete Docker lifecycle again; tasks
are never accepted from an incrementally modified debugging container.

\paragraph{Verifier design.}
Verifiers evaluate observable final-state content and behavior rather than
exact command sequences or intermediate files, allowing alternative correct
solutions to pass. Nondeterministic values, such as timestamps, generated
identifiers, and irrelevant ordering, are normalized when they are not part of
the task requirements. Local services are accessed through deterministic
interfaces, and verifier failures provide localized messages that can be
routed to the corresponding repair agent.

\section{Training and Evaluation Configuration}
\label{app:config}

Table~\ref{tab:config} summarizes the principal training and evaluation
settings. All full-parameter supervised fine-tuning experiments are conducted
with LLaMA-Factory on eight NVIDIA H200 GPUs. The three Qwen3.5 model scales
use the same training recipe and are trained for three epochs.

All evaluations use the Terminus-2 agent scaffold with three attempts per task
and a two-hour timeout per attempt. The FACET fine-tuned models are evaluated
with a maximum context length of 32,768 tokens, while all other models use the
maximum context length supported by their official checkpoints or interfaces.

\begin{table}[t]
    \caption{Principal training and evaluation configurations.}
    \label{tab:config}
    \centering
    \small
    \begin{tabular}{p{0.38\linewidth}p{0.52\linewidth}}
        \toprule
        Setting & Value \\
        \midrule
        \multicolumn{2}{l}{\textbf{Supervised fine-tuning}} \\
        Base models
            & Qwen3.5-4B, Qwen3.5-9B, Qwen3.5-27B \\
        Training data
            & 1.2K complete successful trajectories \\
        Training strategy
            & Full-parameter SFT with BF16 and ZeRO-3 \\
        Epochs / effective batch size
            & 3 / 64 \\
        Learning rate / schedule
            & $1\times10^{-5}$ / cosine with 0.1 warmup ratio \\
        Maximum sequence length
            & 32,768 tokens \\
        \midrule
        \multicolumn{2}{l}{\textbf{Evaluation}} \\
        Benchmark / agent
            & Terminal-Bench 2.1 / Terminus-2 \\
        Attempts / timeout
            & 3 per task / 2 hours per attempt \\
        Temperature
            & 1.0 \\
        Context length
            & FACET models: 32,768 tokens; other models: officially supported maximum \\
        \bottomrule
    \end{tabular}
\end{table}

\section{End-to-End Pipeline Ablation}
\label{app:pipeline-ablation}

We compare three task-construction pipelines using the same 500 accepted
skill-pair inputs. All pipelines use DeepSeek-V4-Pro as the generation model
\citep{deepseekai2026deepseekv4} and produce tasks in the same Harbor format.
The variants differ in scenario construction, artifact organization,
information flow, validation, and repair. This experiment therefore compares
their end-to-end construction designs rather than isolating a single prompt.

\paragraph{Implementation protocol.}
The results are obtained from our implementations under a shared experimental
setting rather than copied from the corresponding papers. We use Codex
\citep{openai2025introducingcodex} to inspect the released repositories,
trace their task-construction workflows, and implement the adapters required
for the comparison. For reproduced pipelines, we preserve the original
prompts, generation order, and artifact-construction logic as closely as
possible. Modifications are limited primarily to accepting the common
skill-pair records, exporting Harbor-compatible task packages, and connecting
the generated tasks to the shared validation and evaluation infrastructure.

\paragraph{Pipeline variants.}
\begin{itemize}
    \item \textbf{Baseline} is a simplified version of our pipeline without
    agentic scenario reconstruction. It directly converts each skill pair into
    a complete task blueprint specifying the intended workflow, artifacts,
    dependencies, interfaces, and success conditions. This static blueprint
    then guides the generation of the environment, instruction, solution, and
    verifier.

    \item \textbf{TW} is our reproduction of the TerminalWorld-style
    construction workflow \citep{chu2026terminalworld}. Codex is used to
    analyze and adapt the released implementation to the shared skill-pair
    inputs. The original prompts and construction logic are retained wherever
    possible, while the necessary input, Harbor-packaging, and validation
    interfaces are added. TW separates environment construction from the
    remaining task artifacts and uses the source skills as references during
    environment and verifier generation.

    \item \textbf{\benchname{} (Ours)} first reconstructs an executable
    scenario from the related skills and preserves the recovered requirements
    through instruction and solution references. It then generates task
    artifacts in stages, explicitly builds and repairs the environment, and
    propagates the shared references across instruction, solution, and
    verifier generation.
\end{itemize}

\paragraph{Evaluation.}
We report both construction yield and the difficulty of the retained tasks.
A complete package contains all required Harbor artifacts, whereas a validated
task additionally requires a buildable environment, a successful oracle
solution, and a verifier that accepts the resulting final state
\citep{harbor2026software}. All validated tasks are subsequently evaluated
using DeepSeek-V4-Pro with Terminus-2
\citep{pi2026dataengineering}. Each task receives three independent attempts
from a clean environment.

\begin{table}[t]
    \caption{End-to-end comparison over 500 common skill-pair inputs.
    Packages denotes complete Harbor task packages, Validated denotes tasks
    passing oracle validation, and Yield is computed over all inputs. P@1 and
    P@3 are evaluated on the tasks retained by each pipeline, and Avg.\ Cmds.\
    is the average number of terminal commands per rollout.}
    \label{tab:pipeline-ablation}
    \centering
    \small
    \begin{tabular*}{\linewidth}{
        @{\extracolsep{\fill}}lrrrrrr@{}
    }
        \toprule
        Pipeline
        & Packages
        & Validated
        & Yield
        & P@1
        & P@3
        & Avg.\ Cmds. \\
        \midrule
        Baseline
        & 437 & 78
        & 15.6\% & 80.8\% & 85.9\% & 12.8 \\
        TW
        & 449 & 139
        & 27.8\% & 58.8\% & 64.0\% & 17.0 \\
        \benchname{} (Ours)
        & 395 & 350
        & 70.0\% & 25.1\% & 33.1\% & 21.5 \\
        \bottomrule
    \end{tabular*}
\end{table}

The comparison yields three main observations:

\begin{itemize}
    \item \textbf{Higher executable-task quality.}
    Although \benchname initially produces fewer complete packages, it
    validates 350 tasks and reaches a 70.0\% end-to-end yield. This exceeds TW
    by 42.2 percentage points and Baseline by 54.4 points. The large gap
    between package generation and validation for the comparison pipelines
    shows that producing all required files does not guarantee consistency
    among the environment, instruction, solution, and verifier. In contrast,
    a substantially larger proportion of \benchname outputs form coherent and
    executable task bundles.

    \item \textbf{Effective validation and repair.}
    Among the 350 accepted \benchname tasks, 182 pass initial validation and
    another 168 are recovered through targeted repair. Execution feedback
    therefore converts many initially inconsistent candidates into valid
    tasks while preserving their underlying scenarios and requirements.

    \item \textbf{The yield gain does not come from easier tasks.}
    The validated \benchname tasks have the lowest P@1 and P@3 and require the
    largest number of terminal commands on average. The pipeline therefore
    produces more valid tasks without reducing them to short or easily solved
    workflows. The retained tasks remain challenging under the same solver
    and agent scaffold.
\end{itemize}

\paragraph{Scope of the comparison.}
TW is a best-effort reproduction under the shared input and evaluation
setting. Although its released prompts and workflow are preserved as closely
as possible, the adapters required for skill-pair inputs, Harbor packaging,
and shared validation may introduce implementation differences from the
original system. Moreover, the pipelines retain different subsets of the
common inputs, so their P@1 and P@3 differences are descriptive rather than a
controlled causal estimate of task difficulty. Construction yield, however,
is measured over the same 500 skill pairs and directly compares how reliably
each pipeline converts the shared source information into valid executable
tasks.

\end{document}